\documentclass[letterpaper]{article} 
\usepackage{aaai23}  
\usepackage{times}  
\usepackage{helvet}  
\usepackage{courier}  
\usepackage[hyphens]{url}  
\usepackage{graphicx} 
\usepackage{natbib}  
\usepackage{caption} 
\usepackage{algorithm}
\usepackage{algorithmic}

\usepackage{newfloat}
\usepackage{listings}
\DeclareCaptionStyle{ruled}{labelfont=normalfont,labelsep=colon,strut=off} 
\floatstyle{ruled}
\newfloat{listing}{tb}{lst}{}
\floatname{listing}{Listing}
\title{Double Graphs Regularized Multi-view Subspace Clustering}
\author{
    Longlong Chen\textsuperscript{\rm 1},
    Yulong Wang\textsuperscript{\rm1 ,}\thanks{Corresponding author.},
    Youheng Liu\textsuperscript{\rm 1},
    Yutao Hu\textsuperscript{\rm 1},
    Libin Wang\textsuperscript{\rm 1}
}
\affiliations{
    \textsuperscript{\rm 1}College of Informatics, Huazhong Agricultural University, Wuhan 430062, China\\

    wangyulong6251@gmail.com
}

\usepackage{bibentry}

\usepackage{cite}
\usepackage{booktabs}
\usepackage{multirow}
\usepackage{amsfonts}
\usepackage{amsmath,amsfonts}
\usepackage{subfigure}
\usepackage{makecell}
\usepackage{mathtools}

\begin{document}

\maketitle

\begin{abstract}
Recent years have witnessed a growing academic interest in multi-view subspace clustering. 
In this paper, we propose a novel Double Graphs Regularized Multi-view Subspace Clustering (DGRMSC) method, which aims to harness both global and local structural information of multi-view data in a unified framework. 
Specifically, DGRMSC firstly learns a latent representation to exploit the global complementary information of multiple views.
Based on the learned latent representation, we learn a self-representation to explore its global cluster structure. 
Further, Double Graphs Regularization (DGR) is performed on both latent representation and self-representation to take advantage of their local manifold structures simultaneously. 
Then, we design an iterative algorithm to solve the optimization problem effectively. 
Extensive experimental results on real-world datasets demonstrate the effectiveness of the proposed method. 
\end{abstract}
\begin{figure*}[!t]
    \centering
    \includegraphics[scale=0.279]{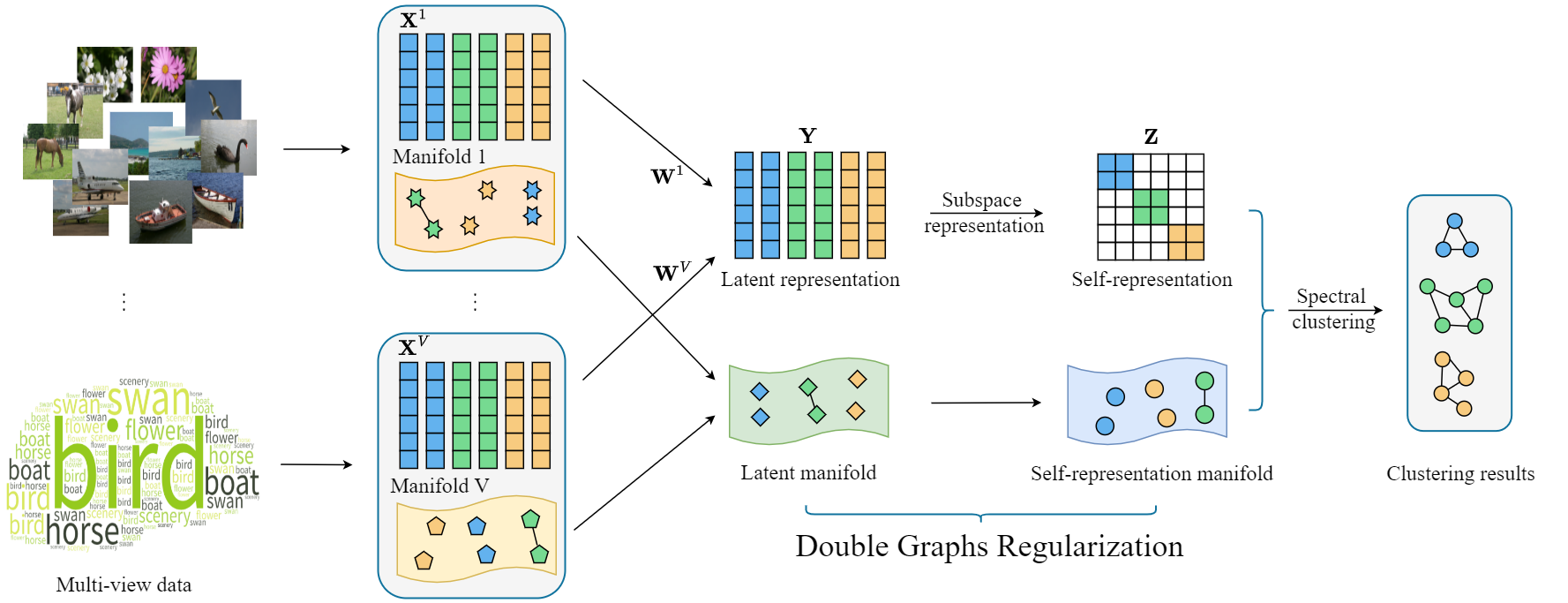}
    \caption{The flowchart of DGRMSC. Given multi-view features $\mathbf{X}^1, \cdots, \mathbf{X}^V$, DGRMSC firstly learns a latent representation $\mathbf{Y}$ to exploit the global complementary information of multiple views. Then, DGRMSC learns its self-representation $\mathbf{Z}$ to explore the global cluster structure. Meanwhile, Double Graphs Regularization (DGR) is performed to preserve the local geometric structures in each manifold throughout the data flow. The points with same color belong to the same class. Two connected points lying in each manifold indicate they are similar or close. }
    \label{Fig0}
\end{figure*}

\section{Introduction}
With the development of information technology, increasing amounts of data are obtained from multiple views. 
For example, an image can be depicted by different features, such as LBP \cite{ojala2002multiresolution}, HOG \cite{dalal2005histograms}, and SIFT \cite{deng2009large}. 
A piece of news can be reported in different languages. 
Meanwhile, in the real world, the data collected is often unlabeled. 
Consequently, it is of significance to study multi-view clustering. 

To process multi-view data, a intuitive approach is to directly use single-view clustering algorithms on the concatenated features of all views. 
Typically, FeatConcate employs standard spectral clustering (SPC) \cite{ng2001spectral} on the concatenated features of all views. 
ConcatePCA firstly applies principle component analysis (PCA) \cite{abdi2010principal} method to extract the low-dimensional representation and then employs SPC to obtain the final results. 
However, this naive strategy neglects the consistency and complementary information among multiple views. 
In the past decades, a large number of multi-view clustering methods have been proposed \cite{zhao2017multi, zhang2018generalized, xie2019multiview, li2021consensus}. 
Among them, Multi-view Subspace Clustering (MSC) is popular since it can effectively capture the global structure and the complementary information of multi-view data. 
For instance, Diversity-induced Multi-view Subspace Clustering (DiMSC) \cite{cao2015diversity} utilizes the Hilbert Schmidt Independence Criterion (HSIC) as a diversity term to explore the complementary of multi-view data. 
Considering both the consistency and the diversity of different views, Luo et al. proposed Consistent and Specific Multi-view Subspace Clustering \cite{luo2018consistent}. 
Lv et al. proposed a partition fusion strategy to enhance the robustness of multi-view clustering \cite{lv2021multi}. 

In real applications, each view is often insufficient and may be contaminated by noise. 
It is worth noting that the methods mentioned above all perform data reconstruction on the original features, which may lead to them being extremely sensitive to the quality of the original data.
To deal with the problem, some multi-view clustering algorithms based on latent subspace have been proposed and achieved sufficiently good performance \cite{zhang2017latent, li2019flexible, yin2020shared, chen2020multi}. 
For example, Latent Multi-view Subspace Clustering (LMSC) \cite{zhang2017latent} clusters data points with latent representation and simultaneously explores underlying complementary information from multiple views. 
Chen et al. proposed a novel Multi-view Clustering in Latent Embedding Space (MCLES) \cite{chen2020multi} method, which simultaneously learns the global structure and the indicator matrix in a unified framework.

However, most latent space based MSC methods do not leverage the manifold structure information within data throughout the data flow. 
In this paper, we propose a novel Double Graphs Regularized Multi-view Subspace Clustering (DGRMSC) method. 
It integrates latent representation learning, self-representation learning, and double graphs regularization into a unified framework. 
The overall flowchart of DGRMSC is shown in Fig. \ref{Fig0}. 
First, a latent representation $\mathbf{Y}$ is learned to exploit the global complementary information of multiple views. 
Then, DGRMSC learns its self-representation $\mathbf{Z}$ to explore the global cluster structure. 
Meanwhile, Double Graphs Regularization (DGR) is performed to preserve the local geometric structure in each manifold throughout the data flow. 
In summary, the main contributions of this paper are as follows: 
\begin{itemize}
    \item We propose a novel Double Graphs Regularized Multi-view Subspace Clustering (DGRMSC) method, which employs both global and local structural information of multi-view data simultaneously in a unified framework.  
    \item To boost the performance of clustering, we propose a novel Double Graphs Regularization (DGR) strategy. It is capable of preserving the local geometric structures in each manifold throughout the data flow. 
    \item We design an iterative algorithm to deal with the optimization problem. Experiments on real-world datasets validate the effectiveness of our method. 
\end{itemize}

\section{The Proposed Approach}
In this section, we will describe the proposed method DGRMSC in detail, including the formulation, optimization, and computational complexity analysis. 

In this paper, vectors and matrices are represented with bold lowercase and uppercase letters respectively.
For a matrix $\mathbf{X}$, $\mathbf{x}_i$ and $x_{ij}$ represent its $i$-th column and $ij$-th element. 
${\Vert \cdot \Vert}_F$ and $ {\Vert \cdot \Vert}_{*}$ denote the Frobenius norm and the nuclear norm of a matrix respectively. 
$\ell_{2,1}$-norm, denoted by ${\Vert \cdot \Vert}_{2,1}$. 
Trace operator of a matrix is denoted by $\operatorname{Tr}\left(\cdot\right)$. 
The key notations used throughout the paper are summarized in Table \ref{tab:table1}.
\begin{table}[!t]
    \centering
    \scalebox{0.89}{
    \begin{tabular}{l|l}
        \Xhline{1pt}
        Notation & Description \\
        \Xhline{0.8pt}
        $\mathbf{X}^v \in \mathbb{R}^{d_v \times n}$                   & Matrix of the $v$-th view \\
        $\mathbf{W}^v \in \mathbb{R}^{d_v \times m}$                   & Mapping matrix of the $v$-th view \\
        $\mathbf{Y} \in \mathbb{R}^{m \times n}$                     & Latent representation \\
        $\mathbf{Z} \in \mathbb{R}^{n \times n}$                     & Self-representation \\
        $\mathbf{A} \in \mathbb{R}^{n \times n}$                     & Affinity matrix \\
        $\mathbf{I}$                     & Identity matrix \\
        $n, V$                           & Number of samples, views \\
        $d_v, m$                         & Dimension of $v$-th view,  latent representation\\
        $d$                              & Total dimension of all views \\
        \Xhline{1pt}
    \end{tabular}
    }
    \caption{Key notations used throughout the paper}
    \label{tab:table1}
\end{table}

\subsection{Latent Represeantation Learning}
Given multi-view data with $n$ samples from $V$ views, we denote the $v$-th view as $\mathbf{X}^v = \left[\mathbf{x}_1^{v},\mathbf{x}_2^{v},\cdots,\mathbf{x}_n^{v}\right] \in \mathbb{R}^{d_v \times n}$, where $v \in \left\{1,2,\cdots,V\right\}$, $d_v$ is the feature dimension of $v$-th view. 
Inspired by \cite{xu2015multi}, it is assumed that each individual view is insufficient for capturing complete information while all the views together can encode complementary information. 
Building on this assumption, we consider all different views $\left\{\mathbf{X}^v\right\}_{v=1}^{V}$ are originated from a shared latent representation $\mathbf{Y}$. 
To be specific, each sample from different views can be reconstructed by their corresponding mapping model $\left\{\mathbf{W}^v\right\}_{v=1}^{V}$ with the shared latent representation $\mathbf{Y} = \left[\mathbf{y}_1, \mathbf{y}_2, \cdots, \mathbf{y}_n\right] \in \mathbb{R}^{m \times n}$, i.e. $\mathbf{x}_i^v = \mathbf{W}^v \mathbf{y}_i, i=1,\cdots,n,v=1,\cdots,V$. 
Considering the noise, the latent representation learning model can be formulated as  
\begin{equation}
    \label{func1}
    \begin{aligned}
        &\mathbf{X} = \mathbf{WY} + \mathbf{E}_L, \text{ s.t. } \mathbf{W}\mathbf{W}^T = \mathbf{I}, \\
        &\mathbf{X} = \left[\begin{array}{c}\mathbf{X}^{1} \\ \cdots \\ \mathbf{X}^{V}\end{array}\right] \text{ and } \mathbf{W} = \left[\begin{array}{c}\mathbf{W}^{1} \\ \cdots \\ \mathbf{W}^{V}\end{array}\right],  
    \end{aligned}
\end{equation} 
where $\mathbf{E}_L \in \mathbb{R}^{d \times n}$ denotes the reconstruction error associated with latent representation learning, where $d = \sum_{v=1}^V d_v$. 
The constraint of $\mathbf{W}$ is to prevent $\mathbf{Y}$ from being pushed arbitrarily close to zero while scaling \cite{zhang2017latent}.
By learning the latent representation, complementary information from multi-view can be effectively explored. 

\subsection{Self-representation Learning}
After acquiring the latent representation $\mathbf{Y}$, a naive approach to obtain the final clustering results is to perform the SPC directly.
However, the restored latent representation $\mathbf{Y}$ ignores the global cluster structure of data which plays a significant role in clustering, and hence may fail in exploring cluster relationship. 
Thus we learn a self-representation $\mathbf{Z} \in \mathbb{R}^{n \times n}$ from the latent representation $\mathbf{Y}$. 
This can be realized by leveraging the self-expression property of samples \cite{ren2020simultaneous}.
It is assumed that each data point can be expressed by a linear combination of other data points. 
Based on the learned latent representation $\mathbf{Y}$, the self-representation learning model is written as 
\begin{equation}
    \label{func2}
    \begin{aligned}
        \mathbf{Y} = \mathbf{YZ} + \mathbf{E}_S,
    \end{aligned}
\end{equation}
where $\mathbf{E}_S \in \mathbb{R}^{m \times n}$ denotes the reconstruction error associated with self-representation learning.   

\subsection{Double Graphs Regularization}
In spite of empirical performance in various applications, both latent space learning and self-representation learning fail to consider the manifold structures in the corresponding manifolds. 
To alleviate this deficiency, one may naturally hope that, if any two multi-view data samples $\{\mathbf{x}_i^v\}_{v=1}^{V}$ and $\{\mathbf{x}_j^v\}_{v=1}^{V}$ are close in the corresponding multi-view manifolds, the corresponding latent representations $\mathbf{y}_i$ and $\mathbf{y}_j$ should also close or similar too. 
To preserve such local manifold structure in the original space, a latent graph regularization term is designed, shown as follows 
\begin{equation}
    \label{func3}
    \begin{aligned}
        \frac{1}{2V} \sum_{v=1}^{V} \sum_{i, j=1}^{n} \left\|\mathbf{y}_{i}-\mathbf{y}_{j}\right\|_{2}^{2} s_{i j}^{v} = \frac{1}{V} \sum_{v=1}^{V} \operatorname{Tr} \left(\mathbf{Y}\mathbf{L}^v\mathbf{Y}^T\right), \\
    \end{aligned}
\end{equation}
where $\mathbf{L}^v = \mathbf{D}^v - \mathbf{S}^v$ is the graph Laplacian matrix of the $v$-th view, and $\mathbf{D}^v$ is a diagonal matrix in which $\mathbf{D}_{ii}^{v} = \sum_{j}^{n} s_{ij}^v$. 
$\mathbf{S}^v$ is the similarity matrix of the $v$-th view.
The $ij$-th element of $\mathbf{S}^v$ is assigned as $s_{i j}^v=\exp \left(-\left\|\mathbf{x}_{i}^v-\mathbf{x}_{j}^v\right\|_{2}^{2} / 2 \sigma^{2}\right)$, where $\sigma$ controls the width of the neighborhoods. 

Analogously, as shown in Fig. \ref{Fig0}, we also want to make the self-representation manifold preserve the local geometric structure of the latent space. 
To this end, we devise the self-representation graph regularization termed as  
\begin{equation}
    \label{func4}
    \begin{aligned}
        \frac{1}{2} \sum_{i, j=1}^{n} \left\|\mathbf{z}_{i}-\mathbf{z}_{j}\right\|_{2}^{2} s_{i j}^{Y} = \operatorname{Tr} \left(\mathbf{Z}\mathbf{L}_{Y}\mathbf{Z}^T\right),  
    \end{aligned}
\end{equation}
where $\mathbf{L}_{Y}$ is the graph Laplacian matrix of latent representation $\mathbf{Y}$. 
Similarly, $s_{i j}^{Y}=\exp \left(-\left\|\mathbf{y}_{i}-\mathbf{y}_{j}\right\|_{2}^{2} / 2 \sigma^{2}\right)$.

Further, we assume that the errors corresponding to the latent representation $\mathbf{Y}$ and the self-representation $\mathbf{Z}$ are sample-specific.
The learned self-representation $\mathbf{Z}$ is low-rank. 
Incorporating Eq. (\ref{func1})-(\ref{func4}) into a unified framework, we have 
\begin{equation}
    \label{func5}
    \begin{aligned}
        \min_{\mathbf{\Theta}} &{\Vert \mathbf{E} \Vert}_{2,1} + \lambda {\Vert \mathbf{Z} \Vert}_{*} + \beta \operatorname{Tr}\left(\mathbf{Y} \mathbf{L} \mathbf{Y}^T\right) + \gamma \operatorname{Tr}\left(\mathbf{Z} \mathbf{L}_{Y} \mathbf{Z}^T\right) \\
        \text{s.t. } &\mathbf{X} = \mathbf{W} \mathbf{Y} + \mathbf{E}_L, \mathbf{Y} = \mathbf{Y} \mathbf{Z} + \mathbf{E}_S, \\
        &\mathbf{E} = \left[\mathbf{E}_L; \mathbf{E}_S\right] \text{ and } \mathbf{W} \mathbf{W}^T = \mathbf{I},
    \end{aligned}
\end{equation}
with $\mathbf{\Theta} \coloneqq \left\{\mathbf{W},\mathbf{Y},\mathbf{Z},\mathbf{E}_L,\mathbf{E}_S\right\}$, where $\lambda > 0$, $\beta > 0$, and $\gamma > 0$ are three non-negative regularization parameters for balancing these four terms and $\mathbf{L} = \frac{1}{V} \sum_{v=1}^{V} \mathbf{L}^v$. 

\section{Optimization}
In this section, we design an algorithm based on the augmented Lagrange multiplier (ALM) with alternating direction minimizing (ADM) framework \cite{lin2011linearized}.  
To make objective function separable, we introduce one auxiliary variable $\mathbf{Q}$ and the objective function can be reformulated as  
\begin{equation}
    \label{func51}
    \begin{aligned}
        \min_{\mathbf{\Theta},\mathbf{Q}} &{\Vert \mathbf{E} \Vert}_{2,1} + \lambda {\Vert \mathbf{Q} \Vert}_{*} + \beta \operatorname{Tr}\left(\mathbf{Y} \mathbf{L} \mathbf{Y}^T\right) + \gamma \operatorname{Tr}\left(\mathbf{Z} \mathbf{L}_{Y} \mathbf{Z}^T\right) \\
        \text{s.t. } &\mathbf{X} = \mathbf{W} \mathbf{Y} + \mathbf{E}_L, \mathbf{Y} = \mathbf{Y} \mathbf{Z} + \mathbf{E}_S, \\
        &\mathbf{E} = \left[\mathbf{E}_L; \mathbf{E}_S\right], \mathbf{W} \mathbf{W}^T = \mathbf{I} \text{ and } \mathbf{Q} = \mathbf{Z}. 
    \end{aligned}
\end{equation}
The Eq. (\ref{func51}) can be solved by minimizing the following problem  
\begin{equation}
    \label{func6}
    \begin{aligned}
        &\mathcal{L} (\mathbf{\Theta}, \mathbf{Q}) = {\Vert \mathbf{E} \Vert}_{2,1} + \lambda {\Vert \mathbf{Q} \Vert}_{*} + \beta \operatorname{Tr}\left(\mathbf{Y} \mathbf{L} \mathbf{Y}^T\right) \\
        &+ \gamma \operatorname{Tr}\left(\mathbf{Z} \mathbf{L}_{Y} \mathbf{Z}^T\right) + \Phi (\mathbf{\Lambda}_{1}, \mathbf{X} - \mathbf{WY} - \mathbf{E}_L) \\
        &+ \Phi (\mathbf{\Lambda}_{2}, \mathbf{Y} - \mathbf{YZ} - \mathbf{E}_S) + \Phi (\mathbf{\Lambda}_{3}, \mathbf{Q} - \mathbf{Z}) \\
        &\text{s.t. } \mathbf{W} \mathbf{W}^T = \mathbf{I},
    \end{aligned}
\end{equation}
with the definition $\Phi(\mathbf{\Lambda}, \mathbf{D}) = \frac{\mu}{2} \left\|\mathbf{D}\right\|_{F}^{2} + \langle\mathbf{\Lambda}, \mathbf{D}\rangle$, where $\langle\cdot,\cdot\rangle$ indicates the matrix inner product and $\mu$ is a positive penalty parameter. 
$\mathbf{\Lambda}_{1}$, $\mathbf{\Lambda}_{2}$ and $\mathbf{\Lambda}_{3}$ are Lagrange multipliers. 
Specifically, we update each variable when fixing the others.
The specific optimization process is as follows.

\textbf{1. }$\mathbf{W}$\textbf{-subproblem:} By fixing the other variables, it is equivalent to solving the following problem 
\begin{equation}
    \label{func7}
    \begin{aligned}
        &\mathbf{W}^{*} = \arg \min_{\mathbf{W}} \frac{\mu}{2} \left\|\left(\mathbf{\Lambda}_{1} / \mu + \mathbf{X} - \mathbf{E}_L\right)^T - \mathbf{Y}^T\mathbf{W}^T\right\|_{F}^{2} \\
        &\text{s.t. } \mathbf{W} \mathbf{W}^T = \mathbf{I}. 
    \end{aligned}
\end{equation}
Based on \cite{huang2013spectral}, the optimal solution to problem (\ref{func7}) is $\mathbf{W}^T=\mathbf{U} \mathbf{V}^T$, where $\mathbf{U}\mathbf{\Sigma}\mathbf{V}^T$ are the Singular Value Decomposition (SVD) of $\mathbf{Y} \left(\mathbf{\Lambda}_{1} / \mu + \mathbf{X} - \mathbf{E}_L\right)^T$. 

\textbf{2. }$\mathbf{Y}$\textbf{-subproblem:} By fixing all the variables except $\mathbf{Y}$, the optimization problem in Eq. (\ref{func6}) is transformed into 
\begin{equation}
    \label{func8}
    \begin{aligned}
        \mathbf{Y}^{*} &= \arg \min_{\mathbf{Y}} \Phi (\mathbf{\Lambda}_{1}, \mathbf{X} - \mathbf{WY} - \mathbf{E}_L) \\
        &+ \Phi (\mathbf{\Lambda}_{2}, \mathbf{Y} - \mathbf{YZ} - \mathbf{E}_S) + \beta \operatorname{Tr}(\mathbf{Y} \mathbf{L} \mathbf{Y}^T). \\
    \end{aligned}
\end{equation}
Taking the derivative of Eq. (\ref{func8}) with respect to $\mathbf{Y}$ and setting it to zero, we have  
\begin{equation}
    \label{func9}
    \begin{aligned}
        &\mathbf{L}_1\mathbf{Y} + \mathbf{Y}\mathbf{R}_1 = \mathbf{C}_1 \\
        \text{with } &\mathbf{L}_1 = \mu \mathbf{W}^T\mathbf{W}, \\
        &\mathbf{R}_1 = \mu (\mathbf{Z} \mathbf{Z}^T - \mathbf{Z} -\mathbf{Z}^T + \mathbf{I}) + \beta (\mathbf{L} + \mathbf{L}^T), \\
        &\mathbf{C}_1 = \mathbf{W}^T \mathbf{\Lambda}_{1} + \mathbf{\Lambda}_{2}(\mathbf{Z}^T - \mathbf{I})\\
        &+ \mu (\mathbf{W}^T\mathbf{X} + \mathbf{E}_S - \mathbf{W}^T\mathbf{E}_L - \mathbf{E}_S\mathbf{Z}^T).
    \end{aligned}
\end{equation} 
The above equation is a standard Sylvester equation which can be solved by utilizing the Bartels-Stewart algorithm \cite{bartels1972solution}. 

\textbf{3. }$\mathbf{Z}$\textbf{-subproblem:} Updating $\mathbf{Z}$ by fixing the other variables is equivalent to solving the following problem 
\begin{equation}
    \label{func10}
    \begin{aligned}
        \mathbf{Z}^{*} &= \arg \min_{\mathbf{Z}} \Phi (\mathbf{\Lambda}_{2}, \mathbf{Y} - \mathbf{YZ} - \mathbf{E}_S) \\
        &+ \Phi (\mathbf{\Lambda}_{3}, \mathbf{Q} - \mathbf{Z}) + \gamma \operatorname{Tr}(\mathbf{Z} \mathbf{L}_{Y} \mathbf{Z}^T). \\
    \end{aligned}
\end{equation}
Taking the derivative of Eq. (\ref{func10}) with respect to $\mathbf{Z}$ and setting it to zero, we have 
\begin{equation}
    \label{func11}
    \begin{aligned}
        &\mathbf{L}_2\mathbf{Z} + \mathbf{Z}\mathbf{R}_2 = \mathbf{C}_2 \\
        \text{with } &\mathbf{L}_2 = \mu (\mathbf{Y}^T\mathbf{Y} + \mathbf{I}), \mathbf{R}_2 = \gamma (\mathbf{L}_{Y} + \mathbf{L}_{Y}^T), \\
        &\mathbf{C}_2 = \mu (\mathbf{Y}^T\mathbf{Y} + \mathbf{Q} - \mathbf{Y}^T\mathbf{E}_S) + \mathbf{\Lambda}_3 + \mathbf{Y}^T\mathbf{\Lambda}_{2}. \\
    \end{aligned}
\end{equation}
Similarly, the above equation can be efficiently solved by the Bartels-Stewart algorithm \cite{bartels1972solution}.
\begin{algorithm}[!t]
    \caption{DGRMSC}
    \label{alg:algorithm1}
    \textbf{Input}: Multi-view data $\mathbf{X}$, hyperparameters $\lambda$, $\beta$ and $\gamma$, and the dimension $m$ of the latent representation $\mathbf{Y}$. \\
    \textbf{Initialize}: $\mathbf{W} = \mathbf{0}$, $\mathbf{E}_L = \mathbf{0}$, $\mathbf{E}_S = \mathbf{0}$, $\mathbf{Q} = \mathbf{Z} = \mathbf{0}$, $\mathbf{\Lambda}_{1} = \mathbf{0}$, $\mathbf{\Lambda}_{2} = \mathbf{0}$, $\mathbf{\Lambda}_{3} = \mathbf{0}$, $\mu = {10}^{-4}$, $\rho = 1.2$, $\epsilon = {10}^{-6}$, $\max_{\mu} = {10}^{6}$; Initialize $\mathbf{Y}$ with random values. 
    \begin{algorithmic}[1]    
        \WHILE{not convergence}
            \STATE Update $\mathbf{W}$, $\mathbf{Y}$, $\mathbf{Z}$, $\mathbf{E}_L$, $\mathbf{E}_S$ and $\mathbf{Q}$ by solving subproblems 1-5; \\
            \STATE Update $\mathbf{\Lambda}_{1}$, $\mathbf{\Lambda}_{2}$ and $\mathbf{\Lambda}_{3}$ according to subproblem 6; \\
            \STATE Update $\mu = \min (\rho \mu , \max_{\mu})$; \\
            \STATE Check the convergence conditions: \\
            ${\Vert \mathbf{X}-\mathbf{W Y}-\mathbf{E}_{L} \Vert}_{\infty} < \epsilon$, ${\Vert \mathbf{Y}-\mathbf{Y Z}-\mathbf{E}_{S} \Vert}_{\infty} < \epsilon$ 
            and ${\Vert \mathbf{Q}-\mathbf{Z} \Vert}_{\infty} < \epsilon$. \\
        \ENDWHILE
    \end{algorithmic}
    \textbf{Output}: $\mathbf{Y}$, $\mathbf{Z}$, $\mathbf{W}$ and $\mathbf{E}$. 
\end{algorithm}

\textbf{4. }$\mathbf{E}$\textbf{-subproblem:} Fix the other variables, the optimization problem in Eq. (\ref{func6}) can be written as 
\begin{equation}
    \label{func12}
    \begin{aligned}
    \mathbf{E}^{*} = \arg \min _{\mathbf{E}} \frac{1}{\mu}\|\mathbf{E}\|_{2,1}+\frac{1}{2}\|\mathbf{E}-\mathbf{G}\|_{F}^{2}, 
    \end{aligned}
\end{equation}
with $\mathbf{G} = \left[\mathbf{X}-\mathbf{WY}+\mathbf{\Lambda}_{1}/\mu; \mathbf{Y}-\mathbf{YZ}+\mathbf{\Lambda}_{2}/\mu\right]$. 
The Lemma 3.2 in \cite{liu2012robust} can be utilized to solve this problem. 

\textbf{5. }$\mathbf{Q}$\textbf{-subproblem:} Fix all the variables except $\mathbf{Q}$, the optimization problem in Eq. (\ref{func6}) becomes 
\begin{equation}
    \label{func13}
    \begin{aligned}
        \mathbf{Q}^{*} = \arg \min _{\mathbf{Q}} \frac{\lambda}{\mu} {\Vert \mathbf{Q} \Vert}_{*} + \frac{1}{2} {\Vert \mathbf{Q}-(\mathbf{Z}-\mathbf{\Lambda}_3/\mu) \Vert}_{F}^{2}.
    \end{aligned} 
\end{equation}
The singular value thresholding operator \cite{cai2010singular} can be used to solve this problem. 

\textbf{6. Updating Multipliers:} The multipliers can be simply updated through 
\begin{equation}
    \label{func15}
    \left\{\begin{aligned}
    \mathbf{\Lambda}_{1} &=\mathbf{\Lambda}_{1}+\mu\left(\mathbf{X}-\mathbf{WY}-\mathbf{E}_{L}\right) \\
    \mathbf{\Lambda}_{2} &=\mathbf{\Lambda}_{2}+\mu\left(\mathbf{Y}-\mathbf{YZ}-\mathbf{E}_{S}\right) \\
    \mathbf{\Lambda}_{3} &=\mathbf{\Lambda}_{3}+\mu(\mathbf{Q}-\mathbf{Z}).
    \end{aligned}\right.
\end{equation}

For clarity, the algorithm of the proposed DGRMSC method is summarized in Algorithm \ref{alg:algorithm1}.
After acquiring the self-representation $\mathbf{Z}$, affinity matrix $\mathbf{A} \in \mathbb{R}^{n \times n}$ is constructed with $\mathbf{A} = \left|\mathbf{Z}\right| + \left|\mathbf{Z}^T\right|$. 
Finally, based on the affinity matrix $\mathbf{A}$, standard spectral clustering algorithm is performed for the final results. 
\begin{table}[!t]
    \centering
    \begin{tabular}{cccc}
        \toprule[1pt]
        \multirow{2}{*}{Dataset} &\multicolumn{3}{c}{Number of} \\ 
                                 & Samples & Views & Clusters \\
            \midrule[0.8pt]
            MSRCV1    & 210     & 6     & 7 \\
            COIL-20   & 1440    & 3     & 20 \\
            Yale      & 165     & 3     & 15 \\
            BBCSport  & 544     & 2     & 5 \\
            100leaves & 1600    & 3     & 100 \\
            BBC       & 685     & 4     & 5 \\   
        \bottomrule[1pt]
    \end{tabular}
    \caption{Characteristics of the datasets}
    \label{tab:table2}
\end{table}
\begin{table*}[!t]
    \centering
    \scalebox{0.746}{
        \begin{tabular}[c]{cccccccc}
            \toprule[1.5pt]
            {Datasets}               & {Methods}                    & {NMI}             & {ACC}             & {F-measure}       & {AR}              & {Recall}          & {Precision} \\
            \midrule[1pt]
            \multirow{13}{*}{MSRCV1} & $\text{SPC}_{\text{BestSV}}$ & $0.6394\pm0.0431$ & $0.7003\pm0.0583$ & $0.6082\pm0.0493$ & $0.5431\pm0.0581$ & $0.6241\pm0.0473$ & $0.5934\pm0.0528$ \\
                                     & $\text{LRR}_{\text{BestSV}}$ & $0.5608\pm0.0080$ & $0.6767\pm0.0063$ & $0.5214\pm0.0056$ & $0.4420\pm0.0068$ & $0.5346\pm0.0046$ & $0.5088\pm0.0074$ \\
                                     & FeatConcate                  & $0.6371\pm0.0455$ & $0.7060\pm0.0749$ & $0.5996\pm0.0606$ & $0.5338\pm0.0707$ & $0.6100\pm0.0603$ & $0.5899\pm0.0623$ \\
                                     & ConcatePCA                   & $0.6468\pm0.0460$ & $0.7330\pm0.0781$ & $0.6165\pm0.0649$ & $0.5540\pm0.0757$ & $0.6221\pm0.0641$ & $0.6111\pm0.0660$ \\
                                     & Co-regularized               & $0.7445\pm0.0205$ & $0.8570\pm0.0412$ & $0.7511\pm0.0285$ & $0.7106\pm0.0339$ & $0.7571\pm0.0208$ & $0.7455\pm0.0360$ \\
                                     & RMSC                         & $0.6810\pm0.0122$ & $0.8140\pm0.0079$ & $0.6808\pm0.0113$ & $0.6288\pm0.0131$ & $0.6881\pm0.0119$ & $0.6737\pm0.0111$ \\
                                     & DiMSC                        & $0.7242\pm0.0137$ & $0.8348\pm0.0094$ & $0.7219\pm0.0122$ & $0.6760\pm0.0143$ & $0.7360\pm0.0130$ & $0.7083\pm0.0127$ \\
                                     & CGD                          & $0.8345\pm0.0016$ & $0.8997\pm0.0017$ & $0.8041\pm0.0028$ & $0.7714\pm0.0033$ & $0.8299\pm0.0018$ & $0.7798\pm0.0037$ \\
                                     & GFSC                         & $0.7843\pm0.0360$ & $0.8567\pm0.0718$ & $0.7619\pm0.0588$ & $0.7218\pm0.0707$ & $0.7860\pm0.0366$ & $0.7408\pm0.0771$ \\
                                     & MCLES                        & $0.7785\pm0.0115$ & $0.8740\pm0.0084$ & $0.7685\pm0.0130$ & $0.7307\pm0.0152$ & $0.7789\pm0.0116$ & $0.7584\pm0.0145$ \\
                                     & MCMLE                        & $0.6986\pm0.0000$ & $0.8095\pm0.0000$ & $0.7035\pm0.0000$ & $0.6545\pm0.0000$ & $0.7195\pm0.0000$ & $0.6881\pm0.0000$ \\
                                     & LMSC                         & $0.6661\pm0.0230$ & $0.7987\pm0.0214$ & $0.6558\pm0.0270$ & $0.5997\pm0.0315$ & $0.6626\pm0.0259$ & $0.6491\pm0.0280$ \\
                                     & \underline{GRMSC}            & $0.8283\pm0.0027$ & $0.9052\pm0.0019$ & $0.8197\pm0.0035$ & $0.7905\pm0.0040$ & $0.8254\pm0.0031$ & $0.8142\pm0.0038$ \\
                                     & \underline{DGRMSC}                      & $\textbf{0.8786}\pm\textbf{0.0049}$ & $\textbf{0.9317}\pm\textbf{0.0023}$ & $\textbf{0.8597}\pm\textbf{0.0037}$ & $\textbf{0.8367}\pm\textbf{0.0042}$ & $\textbf{0.8747}\pm\textbf{0.0046}$ & $\textbf{0.8453}\pm\textbf{0.0029}$ \\
            \midrule[0.8pt]
            \multirow{13}{*}{COIL-20} & $\text{SPC}_{\text{BestSV}}$ & $0.8189\pm0.0166$ & $0.7122\pm0.0377$ & $0.6777\pm0.0365$ & $0.6603\pm0.0388$ & $0.7058\pm0.0275$ & $0.6524\pm0.0463$ \\
                                      & $\text{LRR}_{\text{BestSV}}$ & $0.8216\pm0.0059$ & $0.7429\pm0.0071$ & $0.6962\pm0.0127$ & $0.6801\pm0.0135$ & $0.7126\pm0.0103$ & $0.6806\pm0.0154$ \\
                                      & FeatConcate                  & $0.8287\pm0.0119$ & $0.7321\pm0.0285$ & $0.7045\pm0.0251$ & $0.6887\pm0.0266$ & $0.7254\pm0.0213$ & $0.6852\pm0.0322$ \\
                                      & ConcatePCA                   & $0.8259\pm0.0157$ & $0.7249\pm0.0393$ & $0.6970\pm0.0343$ & $0.6807\pm0.0364$ & $0.7217\pm0.0269$ & $0.6745\pm0.0436$ \\
                                      & Co-regularized               & $0.8196\pm0.0134$ & $0.7217\pm0.0350$ & $0.6870\pm0.0308$ & $0.6702\pm0.0326$ & $0.7057\pm0.0256$ & $0.6694\pm0.0371$ \\
                                      & RMSC                         & $0.8381\pm0.0109$ & $0.7477\pm0.0257$ & $0.7226\pm0.0234$ & $0.7079\pm0.0247$ & $0.7324\pm0.0192$ & $0.7131\pm0.0281$ \\
                                      & DiMSC                        & $0.8333\pm0.0122$ & $0.7566\pm0.0198$ & $0.7159\pm0.0233$ & $0.7009\pm0.0246$ & $0.7280\pm0.0236$ & $0.7043\pm0.0240$ \\
                                      & CGD                          & $0.8779\pm0.0031$ & $0.7900\pm0.0070$ & $0.7645\pm0.0061$ & $0.7513\pm0.0065$ & $0.8295\pm0.0100$ & $0.7091\pm0.0085$ \\
                                      & GFSC                         & $0.8271\pm0.0188$ & $0.6970\pm0.0378$ & $0.6428\pm0.0464$ & $0.6217\pm0.0503$ & $0.7365\pm0.0276$ & $0.5731\pm0.0612$ \\
                                      & MCLES                        & $0.8956\pm0.0001$ & $0.7953\pm0.0003$ & $0.7600\pm0.0004$ & $0.7462\pm0.0005$ & $0.8458\pm0.0001$ & $0.6900\pm0.0007$ \\
                                      & MCMLE                        & $0.8898\pm0.0000$ & $0.6799\pm0.0000$ & $0.6999\pm0.0000$ & $0.6804\pm0.0000$ & $0.7850\pm0.0000$ & $0.5648\pm0.0000$ \\
                                      & LMSC                         & $0.8223\pm0.0105$ & $0.7389\pm0.0195$ & $0.6991\pm0.0191$ & $0.6832\pm0.0202$ & $0.7103\pm0.0188$ & $0.6883\pm0.0212$ \\
                                      & \underline{GRMSC}            & $0.9124\pm0.0041$ & $0.8467\pm0.0082$ & $0.8201\pm0.0088$ & $0.8106\pm0.0093$ & $0.8533\pm0.0079$ & $0.8050\pm0.0110$ \\
                                      & \underline{DGRMSC}           & $\textbf{0.9368}\pm\textbf{0.0002}$ & $\textbf{0.8856}\pm\textbf{0.0157}$ & $\textbf{0.8631}\pm\textbf{0.0098}$ & $\textbf{0.8560}\pm\textbf{0.0103}$ & $\textbf{0.8866}\pm\textbf{0.0002}$ & $\textbf{0.8579}\pm\textbf{0.0157}$ \\
            \midrule[0.8pt]
            \multirow{13}{*}{Yale}    & $\text{SPC}_{\text{BestSV}}$ & $0.6507\pm0.0301$ & $0.6137\pm0.0437$ & $0.4681\pm0.0378$ & $0.4317\pm0.0408$ & $0.4929\pm0.0350$ & $0.4460\pm0.0415$ \\
                                      & $\text{LRR}_{\text{BestSV}}$ & $0.7101\pm0.0110$ & $0.6986\pm0.0158$ & $0.5500\pm0.0176$ & $0.5198\pm0.0189$ & $0.5670\pm0.0166$ & $0.5341\pm0.0194$ \\
                                      & FeatConcate                  & $0.6580\pm0.0310$ & $0.6226\pm0.0455$ & $0.4831\pm0.0407$ & $0.4476\pm0.0437$ & $0.5101\pm0.0402$ & $0.4592\pm0.0426$ \\
                                      & ConcatePCA                   & $0.6502\pm0.0345$ & $0.5986\pm0.0454$ & $0.4668\pm0.0412$ & $0.4297\pm0.0446$ & $0.4999\pm0.0396$ & $0.4384\pm0.0439$ \\
                                      & Co-regularized               & $0.6493\pm0.0386$ & $0.6073\pm0.0624$ & $0.4692\pm0.0536$ & $0.4326\pm0.0580$ & $0.4974\pm0.0494$ & $0.4447\pm0.0582$ \\
                                      & RMSC                         & $0.6797\pm0.0357$ & $0.6511\pm0.0435$ & $0.5188\pm0.0454$ & $0.4865\pm0.0485$ & $0.5344\pm0.0462$ & $0.5041\pm0.0452$ \\
                                      & DiMSC                        & $0.7377\pm0.0160$ & $0.6941\pm0.0268$ & $0.5682\pm0.0252$ & $0.5384\pm0.0271$ & $0.6028\pm0.0244$ & $0.5377\pm0.0290$ \\
                                      & CGD                          & $0.5210\pm0.0270$ & $0.4398\pm0.0300$ & $0.3396\pm0.0276$ & $0.2812\pm0.0324$ & $0.5069\pm0.0165$ & $0.2564\pm0.0294$ \\
                                      & GFSC                         & $0.7006\pm0.0258$ & $0.6596\pm0.0433$ & $0.5142\pm0.0344$ & $0.4796\pm0.0374$ & $0.5628\pm0.0316$ & $0.4738\pm0.0393$ \\
                                      & MCLES                        & $0.7263\pm0.0213$ & $0.6966\pm0.0243$ & $0.5308\pm0.0324$ & $0.4967\pm0.0353$ & $0.5956\pm0.0290$ & $0.4793\pm0.0368$ \\
                                      & MCMLE                        & $0.6882\pm0.0000$ & $0.6303\pm0.0000$ & $0.5041\pm0.0000$ & $0.4685\pm0.0000$ & $0.5588\pm0.0000$ & $0.4592\pm0.0000$ \\
                                      & LMSC                         & $0.7168\pm0.0069$ & $0.6758\pm0.0130$ & $0.5280\pm0.0076$ & $0.4947\pm0.0082$ & $0.5745\pm0.0076$ & $0.4885\pm0.0084$ \\
                                      & \underline{GRMSC}            & $0.7257\pm0.0061$ & $0.7036\pm0.0054$ & $0.5444\pm0.0115$ & $0.5127\pm0.0124$ & $0.5829\pm0.0100$ & $0.5107\pm0.0127$ \\
                                      & \underline{DGRMSC}           & $\textbf{0.7604}\pm\textbf{0.0088}$ & $\textbf{0.7317}\pm\textbf{0.0027}$ & $\textbf{0.6175}\pm\textbf{0.0114}$ & $\textbf{0.5917}\pm\textbf{0.0122}$ & $\textbf{0.7604}\pm\textbf{0.0088}$ & $\textbf{0.5960}\pm\textbf{0.0118}$ \\
            \midrule[0.8pt]
            \multirow{13}{*}{100leaves} & $\text{SPC}_{\text{BestSV}}$ & $0.7818\pm0.0051$ & $0.5693\pm0.0147$ & $0.4599\pm0.0141$ & $0.4545\pm0.0142$ & $0.4907\pm0.0136$ & $0.4328\pm0.0156$ \\
                                        & $\text{LRR}_{\text{BestSV}}$ & $0.7302\pm0.0052$ & $0.5011\pm0.0147$ & $0.3606\pm0.0125$ & $0.3543\pm0.0126$ & $0.3794\pm0.0127$ & $0.3436\pm0.0127$ \\
                                        & FeatConcate                  & $0.9065\pm0.0086$ & $0.7617\pm0.0248$ & $0.7081\pm0.0258$ & $0.7051\pm0.0261$ & $0.7599\pm0.0208$ & $0.6631\pm0.0308$ \\
                                        & ConcatePCA                   & $0.9064\pm0.0089$ & $0.7652\pm0.0227$ & $0.7088\pm0.0247$ & $0.7059\pm0.0249$ & $0.7593\pm0.0242$ & $0.6648\pm0.0267$ \\
                                        & Co-regularized               & $0.9148\pm0.0094$ & $0.7793\pm0.0269$ & $0.7310\pm0.0273$ & $0.7283\pm0.0276$ & $0.7873\pm0.0222$ & $0.6824\pm0.0322$ \\
                                        & RMSC                         & $0.8865\pm0.0049$ & $0.7508\pm0.0130$ & $0.6699\pm0.0132$ & $0.6666\pm0.0134$ & $0.7064\pm0.0132$ & $0.6369\pm0.0145$ \\
                                        & DiMSC                        & $0.8471\pm0.0092$ & $0.6772\pm0.0230$ & $0.5838\pm0.0216$ & $0.5796\pm0.0218$ & $0.6204\pm0.0199$ & $0.5513\pm0.0235$ \\
                                        & CGD                          & $0.9041\pm0.0103$ & $0.7815\pm0.0231$ & $0.6771\pm0.0520$ & $0.6734\pm0.0528$ & $0.8423\pm0.0130$ & $0.5688\pm0.0672$ \\
                                        & GFSC                         & $0.9007\pm0.0079$ & $0.7073\pm0.0222$ & $0.6480\pm0.0310$ & $0.6441\pm0.0314$ & $0.7801\pm0.0157$ & $0.5551\pm0.0397$ \\
                                        & MCLES                        & $0.9275\pm0.0042$ & $0.8071\pm0.0126$ & $0.6945\pm0.0297$ & $0.6911\pm0.0301$ & $0.8306\pm0.0098$ & $0.5978\pm0.0412$ \\
                                        & MCMLE                        & $0.8808\pm0.0000$ & $0.7513\pm0.0000$ & $0.6662\pm0.0000$ & $0.6628\pm0.0000$ & $0.7182\pm0.0000$ & $0.6212\pm0.0000$ \\
                                        & LMSC                         & $0.8720\pm0.0059$ & $0.7333\pm0.0134$ & $0.6383\pm0.0141$ & $0.6347\pm0.0143$ & $0.6819\pm0.0152$ & $0.6001\pm0.0146$ \\
                                        & \underline{GRMSC}            & $0.9586\pm0.0058$ & $0.8771\pm0.0168$ & $0.8526\pm0.0186$ & $0.8511\pm0.0188$ & $0.8941\pm0.0163$ & $0.8149\pm0.0224$ \\
                                        & \underline{DGRMSC}           & $\textbf{0.9657}\pm\textbf{0.0043}$ & $\textbf{0.8996}\pm\textbf{0.0116}$ & $\textbf{0.8756}\pm\textbf{0.0164}$ & $\textbf{0.8744}\pm\textbf{0.0165}$ & $\textbf{0.9190}\pm\textbf{0.0110}$ & $\textbf{0.8363}\pm\textbf{0.0225}$ \\
            \bottomrule[1.5pt]    
        \end{tabular}
    }
    \caption{Performance comparison of clustering on MSRCV1, COIL-20, Yale and 100leaves. The best results are highlighted in bold. The underlined ones are our methods.}
    \label{tab:table3}
\end{table*}

\subsection{Computational Complexity Analysis} 
In this section, the computational complexity analysis will be presented, which consists of constructing similarity matrix and solving Eq. (\ref{func6}). 
Specifically, the similarity matrix is constructed in $k$-nearest neighbors fashion by heat kernel strategy \cite{belkin2001laplacian}, which costs $\mathcal{O} (n^2 d)$. 
Updating $\mathbf{W}$ takes $\mathcal{O} (m^2d + d^3)$. 
The Bartels-Stewart algorithm \cite{bartels1972solution} is used to update $\mathbf{Y}$ and $\mathbf{Z}$, which leads to $\mathcal{O} (m^3)$. 
The complexity of updating $\mathbf{Q}$ is $\mathcal{O} (n^3)$. 
The main complexity of updating $\mathbf{E}$ and the multipliers is the matrix multiplication, which is $\mathcal{O} (mn^2 + mdn)$. 
In summary, the overall computational complexity is $\mathcal{O} (n^2 d) + \mathcal{O} ((m^2d + d^3 + n^3 + 2m^3 + mn^2 + mdn)t)$, where $t$ is the number of iterators. 
Moreover, Algorithm \ref{alg:algorithm1} usually converges within a few steps. 
Thus, under the condition $m \ll d$ the total computational complexity is $\mathcal{O} (d^3+n^3)$. 

\section{Experiments}
In this section, extensive experiments are constructed to evaluate the effectiveness of the proposed method on six real-world multi-view datasets. 

\textbf{Datasets}: In the following experiments, six benchmark datasets are adopted to evaluate the performance of our method, including MSRCV1 \cite{xu2016discriminatively}, COIL-20 \cite{nene1996columbia}, Yale\footnote{http://vision.ucsd.edu/content/yale-face-database}, BBCSport \cite{xia2014robust}, 100leaves\footnote{https://archive.ics.uci.edu/ml/datasets/One-hundred+plant\\+species+leaves+data+set} and BBC\footnote{http://mlg.ucd.ie/datasets/segment.html}. 
The detailed characteristics of the datasets are summarized in Table \ref{tab:table2}. \par
\begin{table*}[!t]
    \centering
    \scalebox{0.688}{
        \begin{tabular}[c]{cccccccc}
            \toprule[1.5pt]
            {Datasets}               & {Methods}                    & {NMI}             & {ACC}             & {F-measure}       & {AR}              & {Recall}          & {Precision} \\
            \midrule[1pt]
            \multirow{13}{*}{BBCSport} & $\text{SPC}_{\text{BestSV}}$ & $0.7901\pm0.0260$ & $0.8627\pm0.0676$ & $0.8487\pm0.0510$ & $0.8025\pm0.0641$ & $0.8423\pm0.0714$ & $0.8566\pm0.0291$ \\
                                       & $\text{LRR}_{\text{BestSV}}$ & $0.8160\pm0.0000$ & $0.9210\pm0.0000$ & $0.8869\pm0.0000$ & $0.8518\pm0.0000$ & $0.8797\pm0.0000$ & $0.8942\pm0.4517$ \\
                                       & FeatConcate                  & $0.7966\pm0.0252$ & $0.8320\pm0.0703$ & $0.8355\pm0.0519$ & $0.7856\pm0.0650$ & $0.8265\pm0.0782$ & $0.8469\pm0.0262$ \\
                                       & ConcatePCA                   & $-$               & $-$               & $-$               & $-$               & $-$               & $-$               \\
                                       & Co-regularized               & $0.8170\pm0.0255$ & $0.8729\pm0.0619$ & $0.8664\pm0.0459$ & $0.8251\pm0.0580$ & $0.8671\pm0.0665$ & $0.8670\pm0.0251$ \\
                                       & RMSC                         & $0.8214\pm0.0002$ & $0.9009\pm0.0006$ & $0.8834\pm0.0002$ & $0.8463\pm0.0003$ & $0.8939\pm0.0000$ & $0.8731\pm0.0004$ \\
                                       & DiMSC                        & $0.8686\pm0.0012$ & $0.9577\pm0.0003$ & $0.9221\pm0.0006$ & $0.8981\pm0.0007$ & $0.9096\pm0.0005$ & $0.9348\pm0.0006$ \\
                                       & CGD                          & $0.9126\pm0.0000$ & $0.9743\pm0.0000$ & $0.9472\pm0.0000$ & $0.9305\pm0.0000$ & $0.9523\pm0.0000$ & $0.9421\pm0.0000$ \\
                                       & GFSC                         & $0.7095\pm0.0582$ & $0.7958\pm0.0786$ & $0.7731\pm0.0691$ & $0.6902\pm0.1034$ & $0.8576\pm0.0301$ & $0.7111\pm0.1044$ \\
                                       & MCLES                        & $0.8268\pm0.0204$ & $0.8775\pm0.0127$ & $0.8784\pm0.0192$ & $0.8380\pm0.0267$ & $0.9179\pm0.0082$ & $0.8424\pm0.0284$ \\
                                       & MCMLE                        & $0.8533\pm0.0000$ & $0.9283\pm0.0000$ & $0.8966\pm0.0000$ & $0.8634\pm0.0000$ & $0.9141\pm0.0000$ & $0.8798\pm0.0000$ \\
                                       & LMSC                         & $0.8200\pm0.0085$ & $0.9254\pm0.0143$ & $0.8900\pm0.0107$ & $0.8558\pm0.0144$ & $0.8846\pm0.0059$ & $0.8956\pm0.0158$ \\
                                       & \underline{GRMSC}                       & $0.9206\pm0.0000$ & $0.9761\pm0.0000$ & $0.9506\pm0.0000$ & $0.9352\pm0.0000$ & $0.9443\pm0.0000$ & $0.9569\pm0.0000$ \\
                                       & \underline{DGRMSC}                      & $\textbf{0.9411}\pm\textbf{0.0000}$ & $\textbf{0.9816}\pm\textbf{0.0000}$ & $\textbf{0.9630}\pm\textbf{0.0000}$ & $\textbf{0.9516}\pm\textbf{0.0000}$ & $\textbf{0.9562}\pm\textbf{0.0000}$ & $\textbf{0.9735}\pm\textbf{0.0000}$ \\
            \midrule[0.8pt]
            \multirow{13}{*}{BBC}       & $\text{SPC}_{\text{BestSV}}$ & $0.4718\pm0.0108$ & $0.6723\pm0.0439$ & $0.5433\pm0.0085$ & $0.4074\pm0.0102$ & $0.5316\pm0.0165$ & $0.5560\pm0.0130$ \\
                                        & $\text{LRR}_{\text{BestSV}}$ & $0.4223\pm0.0020$ & $0.6975\pm0.0012$ & $0.5958\pm0.0020$ & $0.4707\pm0.0025$ & $0.6001\pm0.0024$ & $0.5915\pm0.0016$ \\
                                        & FeatConcate                  & $0.6020\pm0.0151$ & $0.7391\pm0.0559$ & $0.6330\pm0.0117$ & $0.5260\pm0.0153$ & $0.6101\pm0.0137$ & $0.6578\pm0.0146$ \\
                                        & ConcatePCA                   & $-$               & $-$               & $-$               & $-$               & $-$               & $-$               \\
                                        & Co-regularized               & $0.5987\pm0.0129$ & $0.7437\pm0.0547$ & $0.6397\pm0.0104$ & $0.5359\pm0.0156$ & $0.6114\pm0.0252$ & $0.6725\pm0.0256$ \\
                                        & RMSC                         & $0.6189\pm0.0003$ & $0.7885\pm0.0008$ & $0.6690\pm0.0005$ & $0.5758\pm0.0007$ & $0.6291\pm0.0003$ & $0.7142\pm0.0007$ \\
                                        & DiMSC                        & $0.7169\pm0.0000$ & $0.8876\pm0.0000$ & $0.8063\pm0.0000$ & $0.7471\pm0.0000$ & $0.8052\pm0.0000$ & $0.8074\pm0.0000$ \\
                                        & CGD                          & $0.7193\pm0.0007$ & $0.8846\pm0.0003$ & $0.8241\pm0.0004$ & $0.7651\pm0.0005$ & $0.8867\pm0.0001$ & $0.7697\pm0.0006$ \\
                                        & GFSC                         & $0.5589\pm0.0282$ & $0.7157\pm0.0428$ & $0.6412\pm0.0165$ & $0.5003\pm0.0346$ & $0.8039\pm0.0548$ & $0.5388\pm0.0513$ \\
                                        & MCLES                        & $0.6029\pm0.0000$ & $0.7343\pm0.0000$ & $0.6648\pm0.0000$ & $0.5311\pm0.0000$ & $0.8480\pm0.0000$ & $0.5467\pm0.0000$ \\
                                        & MCMLE                        & $0.7465\pm0.0000$ & $0.9022\pm0.0000$ & $0.8274\pm0.0000$ & $0.7722\pm0.0000$ & $0.8554\pm0.0000$ & $0.8012\pm0.0000$ \\
                                        & LMSC                         & $0.6834\pm0.0006$ & $0.8775\pm0.0004$ & $0.7913\pm0.0004$ & $0.7265\pm0.0006$ & $0.8001\pm0.0003$ & $0.7827\pm0.0006$ \\
                                        & \underline{GRMSC}            & $0.7518\pm0.0000$ & $0.8905\pm0.0000$ & $0.8314\pm0.0000$ & $0.7809\pm0.0000$ & $0.8249\pm0.0000$ & $0.8458\pm0.0000$ \\
                                        & \underline{DGRMSC}           & $\textbf{0.8073}\pm\textbf{0.0000}$ & $\textbf{0.9258}\pm\textbf{0.0000}$ & $\textbf{0.8757}\pm\textbf{0.0000}$ & $\textbf{0.8364}\pm\textbf{0.0000}$ & $\textbf{0.8960}\pm\textbf{0.0000}$ & $\textbf{0.8562}\pm\textbf{0.0000}$ \\
            \bottomrule[1.5pt]    
        \end{tabular}
    }
    \caption{Performance comparison of clustering on BBC and BBCSport. The best results are highlighted in bold. The underlined ones are our methods. The clustering results of ConcatePCA are not presented, since the features of both datasets are too sparse to run PCA.}
    \label{tab:table4}
\end{table*}
\textbf{Compared Methods}: Thirteen state-of-the-art methods are compared with the proposed method: $\text{SPC}_\text{BestSV}$ \cite{ng2001spectral}, $\text{LRR}_\text{BestSV}$ \cite{liu2012robust}, FeatConcate, ConcatePCA, Co-regularized \cite{kumar2011co}, RMSC \cite{xia2014robust}, DiMSC \cite{cao2015diversity}, LMSC \cite{zhang2017latent}, CGD \cite{tang2020cgd}, GFSC \cite{kang2020multi}, MCLES \cite{chen2020multi}, and MCMLE \cite{zhong2021improved}. 
Note that $\text{SPC}_\text{BestSV}$ and $\text{LRR}_\text{BestSV}$ indicate the best single-view according to SPC and LRR. 
Further, we report the performance of our method with $\gamma = 0$ as an ablation comparison. 
In this case, only the latent graph regularization term in Eq. (\ref{func3}) is used and the corresponding method is termed as GRMSC (Graph Regularized Multi-view Subspace Clustering).    
\begin{figure*}[!t]
    \centering
    \subfigure[View 1]{
        \label{Fig1.sub.1}
		\includegraphics[width=0.2283\linewidth]{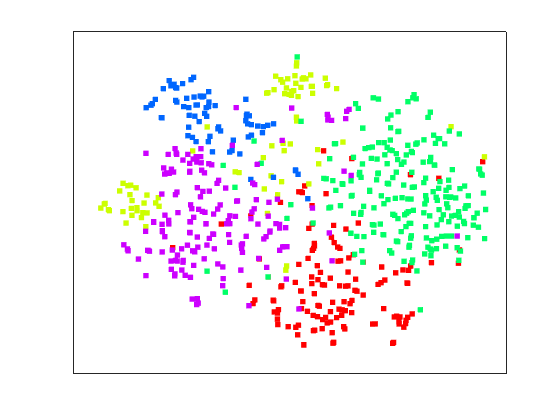}
    }
    \subfigure[View 2]{
        \label{Fig1.sub.2}
		\includegraphics[width=0.2283\linewidth]{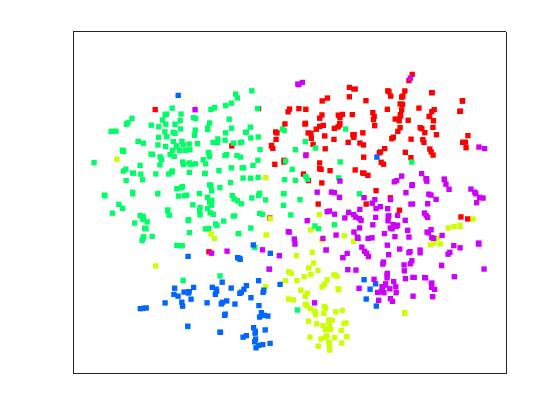}
    }
    \subfigure[View 3]{
        \label{Fig1.sub.3}
		\includegraphics[width=0.2283\linewidth]{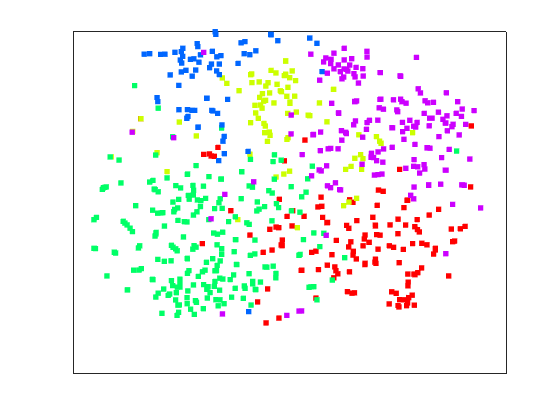}
    }
    \subfigure[View 4]{
        \label{Fig1.sub.4}
		\includegraphics[width=0.223\linewidth]{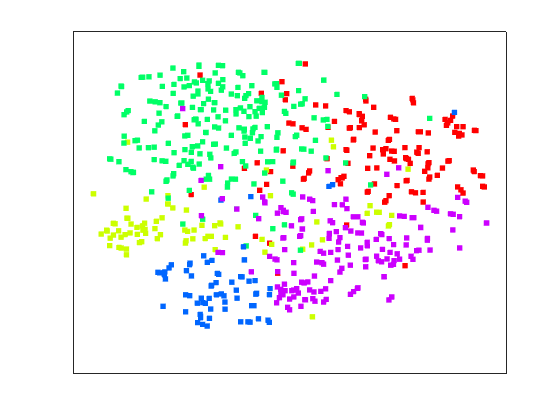}
    }
    \subfigure[LMSC]{
        \label{Fig1.sub.5}
        \includegraphics[width=0.2283\linewidth]{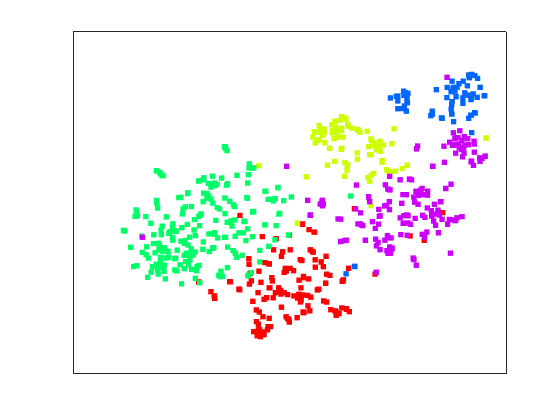}
    }
    \subfigure[MCLES]{
        \label{Fig1.sub.6}
		\includegraphics[width=0.2283\linewidth]{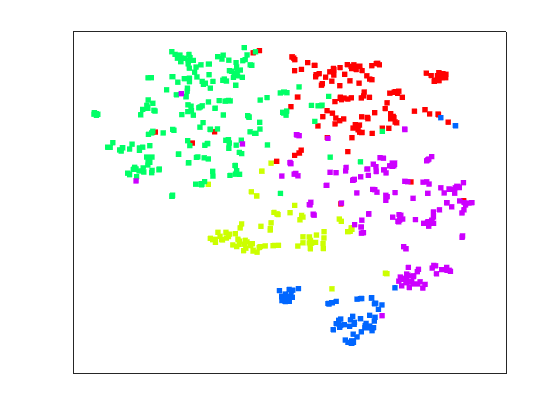}
    }
    \subfigure[GRMSC]{
        \label{Fig1.sub.7}
		\includegraphics[width=0.2283\linewidth]{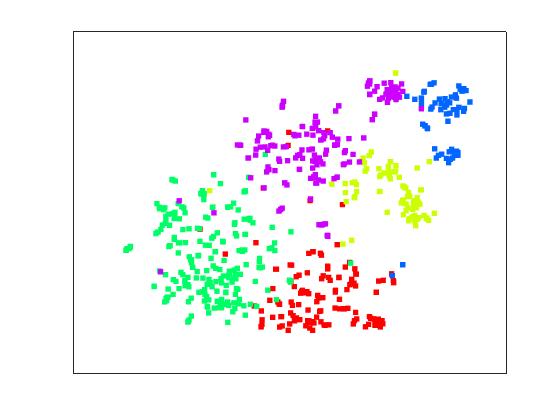}
    }
    \subfigure[DGRMSC]{
        \label{Fig1.sub.8}
		\includegraphics[width=0.2283\linewidth]{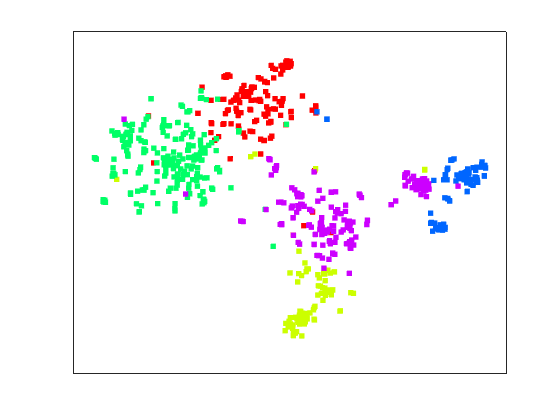}
    }
    \caption{Visualization of different views and different latent representations learned by LMSC, MCLES, GRMSC, and DGRMSC with t-SNE on the BBC dataset.}
    \label{Fig1}
\end{figure*}

\textbf{Evaluation Metrics}: For evaluation metrics, six widely used metrics including F-measure, Precision, Recall, Normalized Mutual Information (NMI), adjusted rand index (AR), and accuracy (ACC) are adopted \cite{li2021consensus}. 
For all metrics, higher values indicate better clustering results. 

\textbf{Parameter Settings}: For the compared methods, the best results are reported through tuning the parameters as suggested in the original paper. 
For our method, the dimension of latent representation $m$ is set to 100 and the parameters $\lambda$, $\beta$, and $\gamma$ are selected from $\{0.001, 0.01, 0.1, 1, 10, 100, 1000\} $ for all datasets. 
The number of nearest neighbors $k$ is set to 5 when constructing $k$-nearest neighbors graph. 
For each method, the mean values and standard deviations of all evaluation metrics are reported after runing 30 times with the optimal parameters.

\subsection{Experimental Results}
\textbf{Comparison Results.} The experimental results on the six datasets are presented in Table \ref{tab:table3} and Table \ref{tab:table4}. 
For the results, the following observations are obtained. 
\begin{itemize}
    \item In most cases, the multi-view methods are always better than the single-view methods, which show that it is of significance to study how to utilize the multi-view information effectively. 
    \item Simply concatenating multiple views and then directly using the single-view methods to deal with it can not achieve effective results, and sometimes even make the results worse. 
    \item The proposed method significantly outperforms the others for all datasets, which validates the effectiveness of the double graphs regularization strategy.   
\end{itemize}
The reasons are as follows. 
Firstly, we cluster on the latent representation rather than operate directly on the original data like other methods, which can effectively reduce the impact of noise and obtain more sufficient information. 
Based on the learned latent representation, the self-representation property can better explore the global cluster structure so that better results can be obtained. 
This can be seen from the results obtained by LMSC and MCLES. 
However, they ignore the manifold structure information within data throughout the data flow, which leads to their results are often suboptimal.
\begin{figure}[!t]
    \centering
    \subfigure[LMSC]{
        \label{Fig2.sub.1}
        \includegraphics[width=0.47\linewidth]{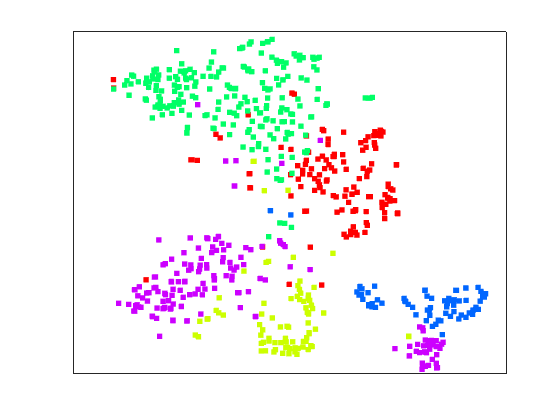}
    }
    \subfigure[MCLES]{
        \label{Fig2.sub.2}
        \includegraphics[width=0.47\linewidth]{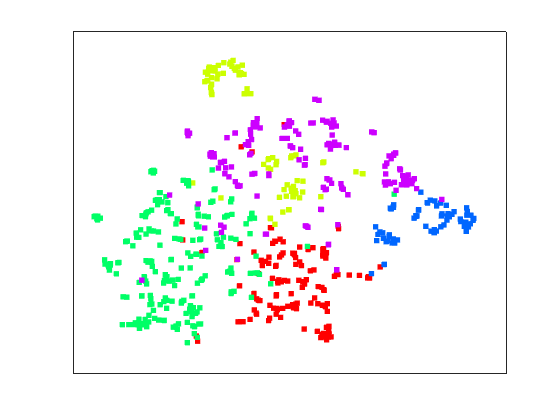}
    }
    \subfigure[GRMSC]{
        \label{Fig2.sub.3}
        \includegraphics[width=0.47\linewidth]{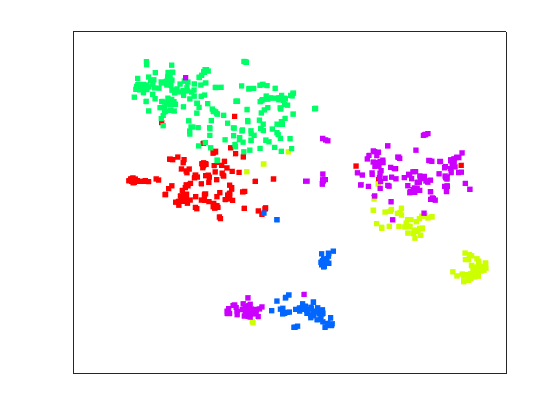}
    }
    \subfigure[DGRMSC]{
        \label{Fig2.sub.4}
        \includegraphics[width=0.47\linewidth]{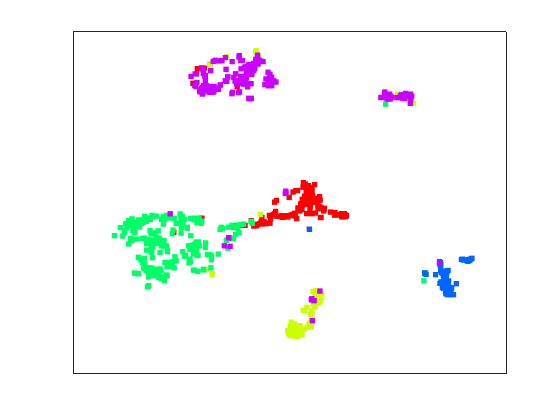}
    }
    \caption{Visualization of different affinity matrices learned by LMSC, MCLES, GRMSC, and DGRMSC on the BBC dataset.}
    \label{Fig2}
\end{figure}
\begin{figure}[!t]
    \centering
    \subfigure[Parameter $\lambda$]{
        \label{Fig3.sub.1}
        \includegraphics[width=0.47\linewidth]{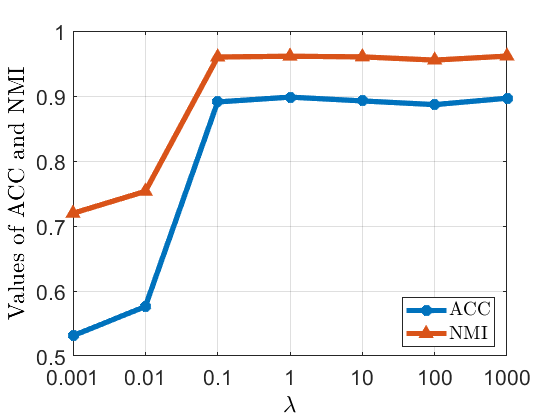}
    }
    \subfigure[Parameter $\beta$]{
        \label{Fig3.sub.2}
        \includegraphics[width=0.47\linewidth]{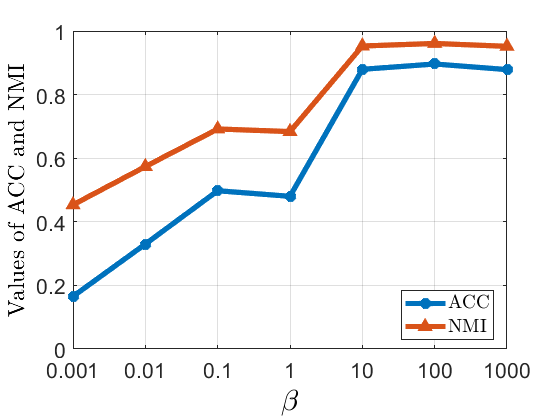}
    }
    \subfigure[Parameter $\gamma$]{
        \label{Fig3.sub.3}
        \includegraphics[width=0.47\linewidth]{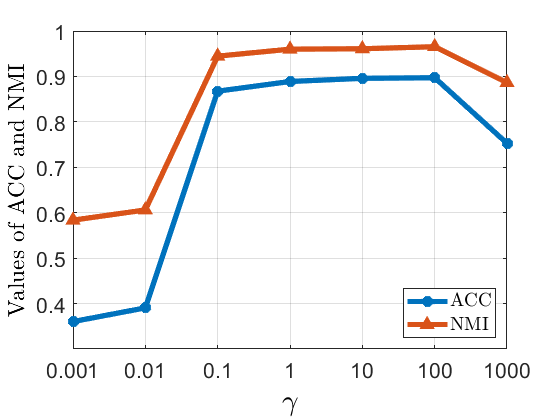}
    }
    \caption{Parameters analysis on $\lambda$, $\beta$, and $\gamma$ in terms of ACC and NMI. When one parameter is analyzed, the other parameters are fixed.}
    \label{Fig3}
\end{figure}
To alleviate the effect, the proposed method performs DGR on both latent representation and self-representation to take advantage of their local manifold structures simultaneously. 
Thus our method obtains better clustering results.  
We also report the performance of our method with $\gamma = 0$ (GRMSC) as an ablation comparison. 
It is observed that GRMSC achieves better performance on most datasets. 
DGRMSC further outperforms GRMSC, which validates the importance of DGR.   

To be more intuitive, we visualize different views and the learned latent representations by different methods with t-Distributed Stochastic Neighbor Embedding (t-SNE) \cite{van2008visualizing} for the dataset BBC as shown in Fig. \ref{Fig1}. 
It is observed that the latent representations learned by different methods are superior to each single view. 
\begin{figure}[!t]
    \centering
    \subfigure[]{
        \label{Fig4.sub.1}
        \includegraphics[width=0.47\linewidth]{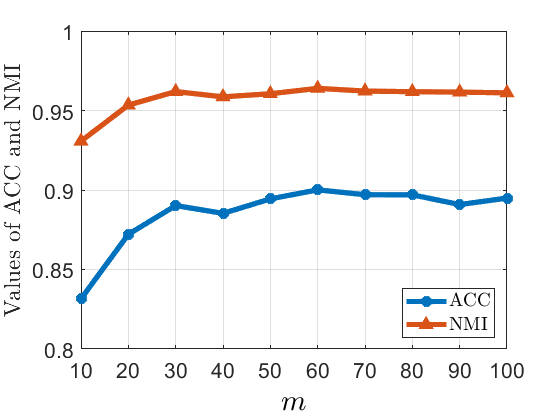}
    }
    \subfigure[]{
        \label{Fig4.sub.2}
        \includegraphics[width=0.47\linewidth]{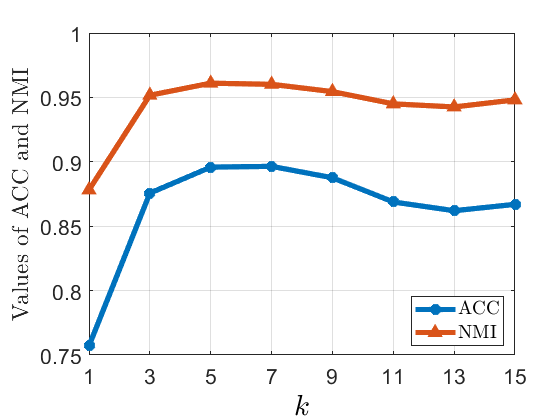}
    }
    \caption{Results of our method when using different dimensionality $m$ and $k$ nearest neighbors value.}
    \label{Fig4}
\end{figure}
Fig. \ref{Fig1}(g)-(h) demonstrates the learned latent representation by our GRMSC and DGRMSC can well explore the complementary information within multi-view data, compared to MCLES and LMSC. 
It is consistent with the performance of clustering in Table \ref{tab:table4}. 
Further, Fig. \ref{Fig2} demonstrates the affinity matrices based clustering learned by LMSC, MCLES, GRMSC, and DGRMSC. 
Obviously, our methods GRMSC and DGRMSC can obtain better results. 
\subsection{Sensitivity Analysis}
Specifically, we take the 100leaves dataset as an example to conduct parameter sensitivity analysis and the results are shown in Fig. \ref{Fig3}. 
It can be seen that our method is relatively parameter insensitive and can obtain good results in a large range. 
Meanwhile, we also analyze the effect of dimensionality $m$ of the latent representation and $k$ nearest neighbors value on the results, as shown in Fig. \ref{Fig4}. 
It can be seen from Fig. \ref{Fig4.sub.1} that our method can achieve good results at lower dimensionality, which further illustrates the effectiveness of latent representation learning. 
Fig. \ref{Fig4.sub.2} shows that our method can achieve a better result when $k = 5$. 
Other datasets have similar properties, so for simplicity, we set $m = 100$ and $k = 5$ in all experiments, respectively. 

\section{Conclusion}
In this paper, a novel Double Graph Regularized Latent Multi-view Subspace Clustering (DGRMSC) method is proposed, which exploits both global and local structural information of multi-view data in a unified framework.
With double graphs regularized strategy, the local geometric information in different manifolds throughout the data flow can be well preserved. 
Accordingly, the learned multi-view latent representation and the affinity matrix based clustering are well improved simultaneously. 
Further, we design an iterative algorithm to solve the optimization problem effectively.  
The experimental results further validate the effectiveness of the proposed method. 

\section{References}

\nobibliography*

\bibentry{ojala2002multiresolution}.\\[.2em]
\bibentry{dalal2005histograms}.\\[.2em]
\bibentry{deng2009large}.\\[.2em]
\bibentry{ng2001spectral}.\\[.2em]
\bibentry{abdi2010principal}.\\[.2em]
\bibentry{cao2015diversity}.\\[.2em]
\bibentry{luo2018consistent}.\\[.2em]
\bibentry{lv2021multi}.\\[.2em]
\bibentry{zhang2017latent}.\\[.2em]
\bibentry{li2019flexible}.\\[.2em]
\bibentry{yin2020shared}.\\[.2em]
\bibentry{chen2020multi}.\\[.2em]
\bibentry{xu2015multi}.\\[.2em]
\bibentry{lin2011linearized}.\\[.2em]
\bibentry{huang2013spectral}.\\[.2em]
\bibentry{bartels1972solution}.\\[.2em]
\bibentry{liu2012robust}.\\[.2em]
\bibentry{cai2010singular}.\\[.2em]
\bibentry{belkin2001laplacian}.\\[.2em]
\bibentry{xu2016discriminatively}.\\[.2em]
\bibentry{xia2014robust}.\\[.2em]
\bibentry{kumar2011co}.\\[.2em]
\bibentry{tang2020cgd}.\\[.2em]
\bibentry{kang2020multi}.\\[.2em]
\bibentry{zhong2021improved}.\\[.2em]
\bibentry{van2008visualizing}.\\[.2em]
\bibentry{li2021consensus}.\\[.2em]
\bibentry{zhao2017multi}.\\[.2em]
\bibentry{zhang2018generalized}.\\[.2em]
\bibentry{xie2019multiview}.


\begin{thebibliography}{32}
\providecommand{\natexlab}[1]{#1}

\bibitem[{Abdi and Williams(2010)}]{abdi2010principal}
Abdi, H.; and Williams, L.~J. 2010.
\newblock Principal component analysis.
\newblock \emph{Wiley interdisciplinary reviews: computational statistics},
  2(4): 433--459.

\bibitem[{Bartels and Stewart(1972)}]{bartels1972solution}
Bartels, R.~H.; and Stewart, G.~W. 1972.
\newblock Solution of the matrix equation AX+ XB= C.
\newblock \emph{Communications of the ACM}, 15(9): 820--826.

\bibitem[{Belkin and Niyogi(2001)}]{belkin2001laplacian}
Belkin, M.; and Niyogi, P. 2001.
\newblock Laplacian eigenmaps and spectral techniques for embedding and
  clustering.
\newblock In \emph{NeurIPS}, volume~14, 585--591.

\bibitem[{Cai, Cand{\`e}s, and Shen(2010)}]{cai2010singular}
Cai, J.-F.; Cand{\`e}s, E.~J.; and Shen, Z. 2010.
\newblock A singular value thresholding algorithm for matrix completion.
\newblock \emph{SIAM Journal on Optimization}, 20(4): 1956--1982.

\bibitem[{Cao et~al.(2015)Cao, Zhang, Fu, Liu, and Zhang}]{cao2015diversity}
Cao, X.; Zhang, C.; Fu, H.; Liu, S.; and Zhang, H. 2015.
\newblock Diversity-induced multi-view subspace clustering.
\newblock In \emph{CVPR}, 586--594.

\bibitem[{Chen et~al.(2020)Chen, Huang, Wang, and Huang}]{chen2020multi}
Chen, M.-S.; Huang, L.; Wang, C.-D.; and Huang, D. 2020.
\newblock Multi-view clustering in latent embedding space.
\newblock In \emph{AAAI}, volume~34, 3513--3520.

\bibitem[{Dalal and Triggs(2005)}]{dalal2005histograms}
Dalal, N.; and Triggs, B. 2005.
\newblock Histograms of oriented gradients for human detection.
\newblock In \emph{CVPR}, volume~1, 886--893.

\bibitem[{Deng et~al.(2009)Deng, Dong, Socher, Li, Li, and
  Fei-Fei}]{deng2009large}
Deng, J.; Dong, W.; Socher, R.; Li, L.-J.; Li, K.; and Fei-Fei, L. 2009.
\newblock ImageNet: A large-scale hierarchical image database.
\newblock In \emph{CVPR}, 248--255.

\bibitem[{Huang, Nie, and Huang(2013)}]{huang2013spectral}
Huang, J.; Nie, F.; and Huang, H. 2013.
\newblock Spectral rotation versus k-means in spectral clustering.
\newblock In \emph{AAAI}, volume~27, 431--437.

\bibitem[{Kang et~al.(2020)Kang, Shi, Huang, Chen, Pu, Zhou, and
  Xu}]{kang2020multi}
Kang, Z.; Shi, G.; Huang, S.; Chen, W.; Pu, X.; Zhou, J.~T.; and Xu, Z. 2020.
\newblock Multi-graph fusion for multi-view spectral clustering.
\newblock \emph{Knowledge-Based Systems}, 189(C): 105102.

\bibitem[{Kumar, Rai, and Daum\'{e}(2011)}]{kumar2011co}
Kumar, A.; Rai, P.; and Daum\'{e}, H. 2011.
\newblock Co-regularized multi-view spectral clustering.
\newblock In \emph{NeurIPS}, volume~24, 1413--1421.

\bibitem[{Li et~al.(2019)Li, Zhang, Hu, Zhu, and Wang}]{li2019flexible}
Li, R.; Zhang, C.; Hu, Q.; Zhu, P.; and Wang, Z. 2019.
\newblock Flexible multi-view representation learning for subspace clustering.
\newblock In \emph{IJCAI}, 2916--2922.

\bibitem[{Li et~al.(2021)Li, Tang, Liu, Zheng, Zhang, and
  Zhu}]{li2021consensus}
Li, Z.; Tang, C.; Liu, X.; Zheng, X.; Zhang, W.; and Zhu, E. 2021.
\newblock Consensus graph learning for multi-view clustering.
\newblock \emph{IEEE Transactions on Multimedia}, 24: 2461--2472.

\bibitem[{Lin, Liu, and Su(2011)}]{lin2011linearized}
Lin, Z.; Liu, R.; and Su, Z. 2011.
\newblock Linearized alternating direction method with adaptive penalty for
  low-rank representation.
\newblock In \emph{NeurIPS}, volume~24, 612--620.

\bibitem[{Liu et~al.(2012)Liu, Lin, Yan, Sun, Yu, and Ma}]{liu2012robust}
Liu, G.; Lin, Z.; Yan, S.; Sun, J.; Yu, Y.; and Ma, Y. 2012.
\newblock Robust recovery of subspace structures by low-rank representation.
\newblock \emph{IEEE Transactions on Pattern Analysis and Machine
  Intelligence}, 35(1): 171--184.

\bibitem[{Luo et~al.(2018)Luo, Zhang, Zhang, and Cao}]{luo2018consistent}
Luo, S.; Zhang, C.; Zhang, W.; and Cao, X. 2018.
\newblock Consistent and specific multi-view subspace clustering.
\newblock In \emph{AAAI}, volume~32, 3730--3737.

\bibitem[{Lv et~al.(2021)Lv, Kang, Wang, Ji, and Xu}]{lv2021multi}
Lv, J.; Kang, Z.; Wang, B.; Ji, L.; and Xu, Z. 2021.
\newblock Multi-view subspace clustering via partition fusion.
\newblock \emph{Information Sciences}, 560: 410--423.

\bibitem[{Nene, Nayar, and Murase(1996)}]{nene1996columbia}
Nene, S.~A.; Nayar, S.~K.; and Murase, H. 1996.
\newblock Columbia object image library (coil-20).

\bibitem[{Ng, Jordan, and Weiss(2001)}]{ng2001spectral}
Ng, A.; Jordan, M.; and Weiss, Y. 2001.
\newblock On spectral clustering: Analysis and an algorithm.
\newblock In \emph{NeurIPS}, 849--856.

\bibitem[{Ojala, Pietikainen, and Maenpaa(2002)}]{ojala2002multiresolution}
Ojala, T.; Pietikainen, M.; and Maenpaa, T. 2002.
\newblock Multiresolution gray-scale and rotation invariant texture
  classification with local binary patterns.
\newblock \emph{IEEE Transactions on Pattern Analysis and Machine
  Intelligence}, 24(7): 971--987.

\bibitem[{Ren and Sun(2020)}]{ren2020simultaneous}
Ren, Z.; and Sun, Q. 2020.
\newblock Simultaneous global and local graph structure preserving for multiple
  kernel clustering.
\newblock \emph{IEEE Transactions on Neural Networks and Learning Systems},
  32(5): 1839--1851.

\bibitem[{Tang et~al.(2020)Tang, Liu, Zhu, Zhu, Luo, Wang, and
  Gao}]{tang2020cgd}
Tang, C.; Liu, X.; Zhu, X.; Zhu, E.; Luo, Z.; Wang, L.; and Gao, W. 2020.
\newblock CGD: Multi-view clustering via cross-view graph diffusion.
\newblock In \emph{AAAI}, volume~34, 5924--5931.

\bibitem[{Van~der Maaten and Hinton(2008)}]{van2008visualizing}
Van~der Maaten, L.; and Hinton, G. 2008.
\newblock Visualizing data using t-SNE.
\newblock \emph{Journal of Machine Learning Research}, 9(11): 2579--2605.

\bibitem[{Xia et~al.(2014)Xia, Pan, Du, and Yin}]{xia2014robust}
Xia, R.; Pan, Y.; Du, L.; and Yin, J. 2014.
\newblock Robust multi-view spectral clustering via low-rank and sparse
  decomposition.
\newblock In \emph{AAAI}, volume~28, 2149--2155.

\bibitem[{Xie et~al.(2019)Xie, Zhang, Gao, Han, Xiao, and
  Gao}]{xie2019multiview}
Xie, D.; Zhang, X.; Gao, Q.; Han, J.; Xiao, S.; and Gao, X. 2019.
\newblock Multiview clustering by joint latent representation and similarity
  learning.
\newblock \emph{IEEE Transactions on Cybernetics}, 50(11): 4848--4854.

\bibitem[{Xu, Tao, and Xu(2015)}]{xu2015multi}
Xu, C.; Tao, D.; and Xu, C. 2015.
\newblock Multi-view intact space learning.
\newblock \emph{IEEE Transactions on Pattern Analysis and Machine
  Intelligence}, 37(12): 2531--2544.

\bibitem[{Xu, Han, and Nie(2016)}]{xu2016discriminatively}
Xu, J.; Han, J.; and Nie, F. 2016.
\newblock Discriminatively embedded k-means for multi-view clustering.
\newblock In \emph{CVPR}, 5356--5364.

\bibitem[{Yin, Huang, and Gao(2020)}]{yin2020shared}
Yin, M.; Huang, W.; and Gao, J. 2020.
\newblock Shared generative latent representation learning for multi-view
  clustering.
\newblock In \emph{AAAI}, volume~34, 6688--6695.

\bibitem[{Zhang et~al.(2018)Zhang, Fu, Hu, Cao, Xie, Tao, and
  Xu}]{zhang2018generalized}
Zhang, C.; Fu, H.; Hu, Q.; Cao, X.; Xie, Y.; Tao, D.; and Xu, D. 2018.
\newblock Generalized latent multi-view subspace clustering.
\newblock \emph{IEEE Transactions on Pattern Analysis and Machine
  Intelligence}, 42(1): 86--99.

\bibitem[{Zhang et~al.(2017)Zhang, Hu, Fu, Zhu, and Cao}]{zhang2017latent}
Zhang, C.; Hu, Q.; Fu, H.; Zhu, P.; and Cao, X. 2017.
\newblock Latent multi-view subspace clustering.
\newblock In \emph{CVPR}, 4279--4287.

\bibitem[{Zhao, Ding, and Fu(2017)}]{zhao2017multi}
Zhao, H.; Ding, Z.; and Fu, Y. 2017.
\newblock Multi-view clustering via deep matrix factorization.
\newblock In \emph{AAAI}, 2921–--2927.

\bibitem[{Zhong and Pun(2021)}]{zhong2021improved}
Zhong, G.; and Pun, C.-M. 2021.
\newblock Improved Normalized Cut for Multi-view Clustering.
\newblock \emph{IEEE Transactions on Pattern Analysis and Machine
  Intelligence}.

\end{thebibliography}
\nobibliography{aaai23}




\end{document}